\documentclass[10pt,twocolumn,letterpaper]{article}

\usepackage{cvpr}
\usepackage{times}
\usepackage{epsfig}
\usepackage{graphicx}
\usepackage{subcaption}
\usepackage{amsmath}
\usepackage{amssymb}
\usepackage{multirow}
\usepackage{caption}

\usepackage[breaklinks=true,bookmarks=false]{hyperref}

\newcommand{\myparagraph}[1]{\vspace{6pt}\noindent{\bf #1}}

\def\mthd{\texttt{f-xGAN}\xspace}

\def\argmax{\mathop{\rm argmax}	}

\def\1{\mathds{1}}

\cvprfinalcopy 

\def\cvprPaperID{2709} 

\ifcvprfinal\fi

\graphicspath{{./figures/}}

\begin{document}

\title{ Feature Generating Networks for Zero-Shot Learning}

 \author{
 Yongqin Xian$^{1}$ \hspace{4mm} Tobias Lorenz$^{1}$ \hspace{4mm} Bernt Schiele$^{1}$ \hspace{4mm} Zeynep Akata$^{1,2}$\vspace{4mm} \\ 
  \begin{tabular}{cc}
  $^{1}$Max Planck Institute for Informatics & $^{2}$Amsterdam Machine Learning Lab \\ Saarland Informatics Campus & University of Amsterdam 
 \end{tabular}
 }
 

\maketitle

\begin{abstract}
Suffering from the extreme training data imbalance between seen and unseen classes,  most of existing state-of-the-art approaches fail to achieve satisfactory results for the challenging generalized zero-shot learning task. 
To circumvent the need for labeled examples of unseen classes, we propose a novel generative adversarial network~(GAN)
that synthesizes CNN features conditioned on class-level semantic information,
offering a shortcut directly from a semantic descriptor of a class to a class-conditional feature distribution.
Our proposed approach, pairing a Wasserstein GAN with a classification loss, is able to generate sufficiently discriminative CNN features to train softmax classifiers or any multimodal embedding method. Our experimental results demonstrate a significant boost in accuracy over the state of the art on five challenging datasets -- 
CUB, FLO, SUN, AWA and ImageNet -- 
in both the zero-shot learning and generalized zero-shot learning settings.
\end{abstract}



\section{Introduction}



Deep learning has allowed to push performance considerably across a wide range of computer vision and machine learning tasks. However, almost always, deep learning requires large amounts of training data which we are lacking in many practical scenarios, e.g. it is impractical to annotate all the concepts that surround us, and have enough of those annotated samples to train a deep network.
Therefore, training data generation has become a hot research topic~\cite{chawla2002smote,GPMXWDOCB14,CK17,RAYLSL16,han2017stackgan,improvedgan}. Generative Adversarial Networks~\cite{GPMXWDOCB14} are particularly appealing as they allow generating realistic and sharp images conditioned, for instance, on object categories~\cite{RAYLSL16,han2017stackgan}. However, they do not yet generate images of sufficient quality to train deep learning architectures as demonstrated by our experimental results.

In this work, we are focusing on arguably the most extreme case of lacking data, namely zero-shot learning~\cite{LNH13,XSA17,CCGS16b}, where the task is to learn to classify when \textit{no} labeled examples of certain classes are available during training. We argue that this scenario is a great testbed for evaluating the robustness and generalization of generative models. In particular, if the generator learns discriminative visual data with enough variation, the generated data should be useful for supervised learning. Hence, one contribution of our paper is a comparison of various existing GAN-models and another competing generative model, i.e. GMMN, for visual feature generation. In particular, we look into both zero-shot learning (ZSL) where the test time search space is restricted to unseen class labels and generalized zero-shot learning (GZSL) for being a more realistic scenario as at test time the classifier has to decide between both seen and unseen class labels. In this context, we propose a novel GAN-method -- namely \texttt{f-CLSWGAN} that generates features instead of images and is trained with a novel loss improving over alternative GAN-models.


\begin{figure}[t]
	\centering
        \includegraphics[width=\columnwidth]{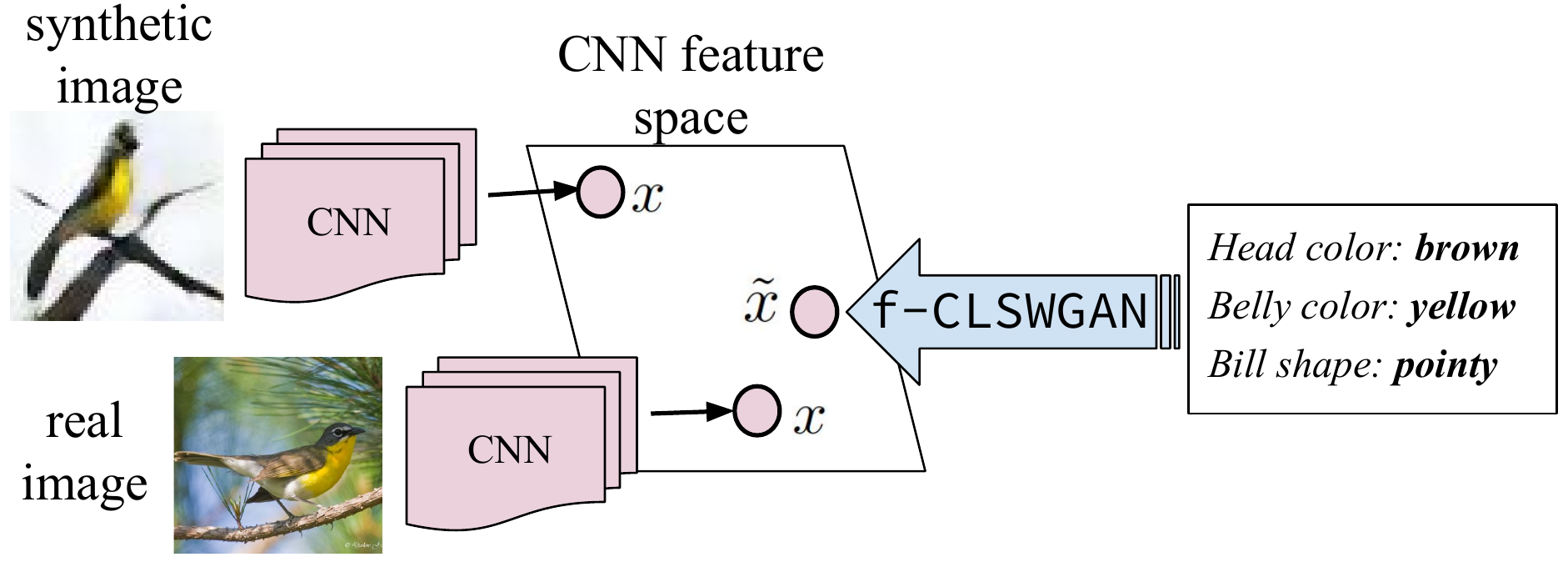}
	\caption{CNN features can be extracted from: 1) real images, however in zero-shot learning we do not have access to any real images of unseen classes, 2) synthetic images, however they are not accurate enough to improve image classification performance. We tackle both of these problems and propose a novel attribute conditional feature generating adversarial network formulation, i.e. \texttt{f-CLSWGAN}, to generate CNN features of unseen classes.}
	\label{fig:teaser}
\end{figure}

We summarize our contributions as follows.
(1) We propose a novel conditional generative model~\texttt{f-CLSWGAN} that synthesizes CNN features of unseen classes by optimizing the Wasserstein distance regularized by a classification loss. 
(2) Across five datasets with varying granularity and sizes, we consistently improve upon the state of the art in both the ZSL and GZSL settings. We demonstrate a practical application for adversarial training and propose GZSL as a proxy task to evaluate the performance of generative models.
(3) Our model is generalizable to different deep CNN features, e.g. extracted from GoogleNet or ResNet, and may use different class-level auxiliary information, e.g. sentence, attribute, and word2vec embeddings. 

\section{Related work}

In this section we review some recent relevant literature on Generative Adversarial Networks, Zero-Shot Learning (ZSL) and Generalized Zero-Shot (GZSL) Learning.

\myparagraph{Generative Adversarial Network.} GAN~\cite{GPMXWDOCB14}
was originally proposed as a means of learning a generative model
which captures an arbitrary data distribution,
such as images, from a particular domain.
The input to a generator network is a ``noise'' vector $z$ drawn from a latent distribution, such as a multivariate Gaussian.
DCGAN~\cite{RMC16} extends GAN by leveraging deep convolution neural networks and providing best practices for GAN training. \cite{WG16} improves DCGAN by factorizing the image generation process into style and structure networks, InfoGAN~\cite{infogan} extends GAN by additionally maximizing the mutual information between interpretable latent variables and the generator distribution. 
GAN has also been extended to a  conditional GAN by feeding the class label~\cite{conditionalgans}, sentence descriptions~\cite{RAMSSL16,RAYLSL16,han2017stackgan}, into both the generator and discriminator. The theory of GAN is recently investigated in \cite{arjovsky2017towards,arjovsky2017wasserstein, gulrajani2017improved}, where they show that the Jenson-Shannon divergence optimized by the original GAN leads to instability issues. To cure the unstable training issues of GANs, \cite{arjovsky2017wasserstein} proposes Wasserstein-GAN~(WGAN), which optimizes an efficient approximation of the Wasserstein distance. While WGAN attains better theoretical properties than the original GAN, it still suffers from vanishing and exploding gradient problems due to weight clipping to enforce the 1-Lipschitz constraint on the discriminator. Hence, \cite{gulrajani2017improved} proposes an improved version of WGAN enforcing the Lipschitz constraint through gradient penalty. Although those papers have demonstrated realistic looking images, they have not applied this idea to image feature generation. 


In this paper, we empirically show that images generated by the state-of-the-art GAN~\cite{gulrajani2017improved} are not ready to be used as training data for learning a classifier. Hence, we propose a novel GAN architecture to directly generate CNN features that can be used to train a discriminative classifier for zero-shot learning. Combining the powerful WGAN~\cite{gulrajani2017improved} loss and a classification loss which enforces the generated features to be discriminative, our proposed GAN architecture improves the original GAN~\cite{GPMXWDOCB14} by a large margin and has an edge over WGAN~\cite{gulrajani2017improved} thanks to our regularizer. 

\myparagraph{ZSL and GZSL.} In the zero-shot learning setting,
the set of classes seen during training and evaluated during test are disjoint~\cite{HEEY15,LNH13,LEB08,RSS11,YA10}.
As supervised learning methods can not be employed for this task, \cite{LNH13,RSS11}
proposed to solve it by solving related sub-problems. \cite{ZV15,NMBSSFCD14,CCGS16}
learn unseen classes as a mixture of seen class proportions, and \cite{APHS15,ARWLS15,FCSBDRM13,SGMN13,XASNHS16,RT15,FS16,QLSH16,AMFS16,BHJ16,FXKG15,KXFG15} learn a compatibility between images and classes. On the other hand, instead of using only labeled data, \cite{FHXFG15, MES13, LGS15} leverage  unlabeled data from unseen classes in the transductive setting. While zero-shot learning has attracted a lot of attention, there has been little work~\cite{SGMN13,CCGS16b} in the more realistic generalized zero-shot learning setting, where both seen and unseen classes appear at test time. 

In this paper, we propose to tackle generalized zero-shot learning by generating CNN features for unseen classes via a novel GAN model. Our work is different from~\cite{HG16} because they generate additional examples for data-starved classes from feature vectors alone, which is unimodal and do not generalize to unseen classes. Our work is closer to~\cite{BHJ17} in which they generate features via GMMN~\cite{li2015generative}. 
Hence, we directly compare with them on the latest zero-shot learning benchmark~\cite{XSA17} and show that WGAN~\cite{arjovsky2017wasserstein} coupled with our proposed classification loss can further improve GMMN in feature generation on most datasets for both ZSL and GZSL tasks.

\section{Feature Generation \& Classification in ZSL}
\label{sec:model}

Existing ZSL models only see labeled data from seen classes during training biasing the predictions to seen classes. The main insight of our proposed model is that by feeding additional synthetic CNN features of unseen classes, the learned classifier will also explore the embedding space of unseen classes. Hence, the key to our approach is the ability to generate semantically rich CNN feature distributions conditioned on a class specific semantic vector e.g. attributes, without access to any images of that class. This alleviates the imbalance between seen and unseen classes, as there is no limit to the number of synthetic CNN features that our model can generate. It also allows to directly train a discriminative classifier, i.e. Softmax classifier, even for unseen classes. 

We begin by defining the problem of our interest. Let  $\mathcal{S} = \{ (x, y, c(y)) | x \in \mathcal{X}, y \in \mathcal{Y}^{s}, c(y) \in \mathcal{C} \}$ where $\mathcal{S}$ stands for the training data of seen classes, $x\in \mathbb{R}^{d_x}$ is the CNN features, $y$ denotes the class label in $\mathcal{Y}^{s}=\{y_1,\ldots, y_K\}$ consisting of K discrete seen classes, and $c(y)\in \mathbb{R}^{d_c}$ is the class embedding, e.g. attributes, of class $y$ that models the semantic relationship between classes. 
In addition, we have a disjoint class label set $\mathcal{Y}^{u}=\{u_1,\ldots, u_L\}$ of unseen classes, whose class embedding set $\mathcal{U}=\{(u, c(u)) | u \in \mathcal{Y}^{u}, c(u)\in \mathcal{C}\}$ is available but images and image features are missing. Given $\mathcal{S}$ and $\mathcal{U}$, the task of ZSL is to learn a classifier $f_{zsl}:\mathcal{X}\rightarrow \mathcal{Y}^{u}$ and in GZSL we learn a classifier $f_{gzsl}:\mathcal{X}\rightarrow \mathcal{Y}^{s} \cup \mathcal{Y}^{u}$.

\subsection{Feature Generation}
\label{sec:GAN}
In this section, we begin our discussion with Generative Adversarial Networks (GAN)~\cite{GPMXWDOCB14} for it being the basis of our model. GAN consists of a generative network $G$ and a discriminative network $D$ that compete in a two player minimax game. In the context of generating image pixels, $D$ tries to accurately distinguish real images from generated images, while $G$ tries to fool the discriminator by generating images that are mistakable for real. Following~\cite{conditionalgans}, we extend GAN to  conditional GAN by including a conditional variable to both $G$ and $D$. In the following we give the details of the conditional GAN variants that we develop. Our novelty lies in that we develop three conditional GAN variants, i.e. \texttt{f-GAN}, \texttt{f-WGAN} and \texttt{f-CLSWGAN}, to generate image features rather than image pixels. It is worth noting that our models are only trained with seen class data $\mathcal{S}$ but can also generate image features of unseen classes.




\myparagraph{\texttt{f-GAN}.} Given the train data $\mathcal{S}$ of seen classes, we aim to learn a conditional generator $G:\mathcal{Z} \times \mathcal{C}\rightarrow \mathcal{X}$, which takes random Gaussian noise $z\in \mathcal{Z} \subset \mathbb{R}^{d_z}$ and class embedding $c(y) \in \mathcal{C}$ as its inputs, and outputs a CNN image feature $\tilde{x} \in \mathcal{X}$ of class $y$. Once the generator $G$ learns to generate CNN features of real images, i.e. $x$, conditioned on the seen class embedding $c(y) \in \mathcal{Y}^s$, it can also generate $\tilde{x}$ of any unseen class $u$ via its class embedding $c(u)$. 
 Our feature generator \texttt{f-GAN} is learned by optimizing the following objective,
%
\begin{align}
\label{eq:gan}
\min_G \max_D \mathcal{L}_{GAN} =& E[\log D(x,c(y))] + \\
        & E[\log \left(1-D(\tilde{x},c(y))\right)],  \nonumber
\end{align}
with $\tilde{x}=G(z, c(y))$. The discriminator $D: \mathcal{X} \times \mathcal{C} \rightarrow [0, 1]$ is a multi-layer perceptron with a sigmoid function as the last layer. While $D$ tries to maximize the loss, $G$ tries to minimizes it. Although GAN has been shown to capture complex data distributions, e.g. pixel images, they are notoriously difficult to train~\cite{arjovsky2017towards}.

\myparagraph{\texttt{f-WGAN}.} 
We extend the improved WGAN~\cite{gulrajani2017improved} 
to a conditional WGAN by integrating the class embedding $c(y)$ to both the generator and the discriminator. The loss is, 
\begin{align}
\label{eq:wgan}
\mathcal{L}_{WGAN} =& E[D(x,c(y))] - E[D(\tilde{x},c(y))] -  \\ 
                   & \lambda E[\left(||\nabla_{\hat{x}} D(\hat{x},c(y))||_2 - 1\right)^2], \nonumber
\end{align}
where $\tilde{x}=G(z,c(y))$, $\hat{x}=\alpha x + (1-\alpha) \tilde{x}$ with $\alpha \sim U(0,1)$, and $\lambda$ is the penalty coefficient. In contrast to the GAN, the discriminative network here is defined as $D: \mathcal{X} \times \mathcal{C} \rightarrow \mathbb{R}$, which eliminates the sigmoid layer and outputs a real value. The log in Equation~\ref{eq:gan} is also removed since we are not optimizing the log likelihood. Instead, the first two terms in Equation~\ref{eq:wgan} approximate the Wasserstein distance, and the third term is the gradient penalty which enforces the gradient of D to have unit norm along the straight line between pairs of real and generated points. Again, we solve a minmax optimization problem, 
\begin{align}
 \min_G \max_D \mathcal{L}_{WGAN}
\end{align}

\begin{figure}[t]
	\centering
        \includegraphics[width=\columnwidth,trim=0 0 0 0,clip]{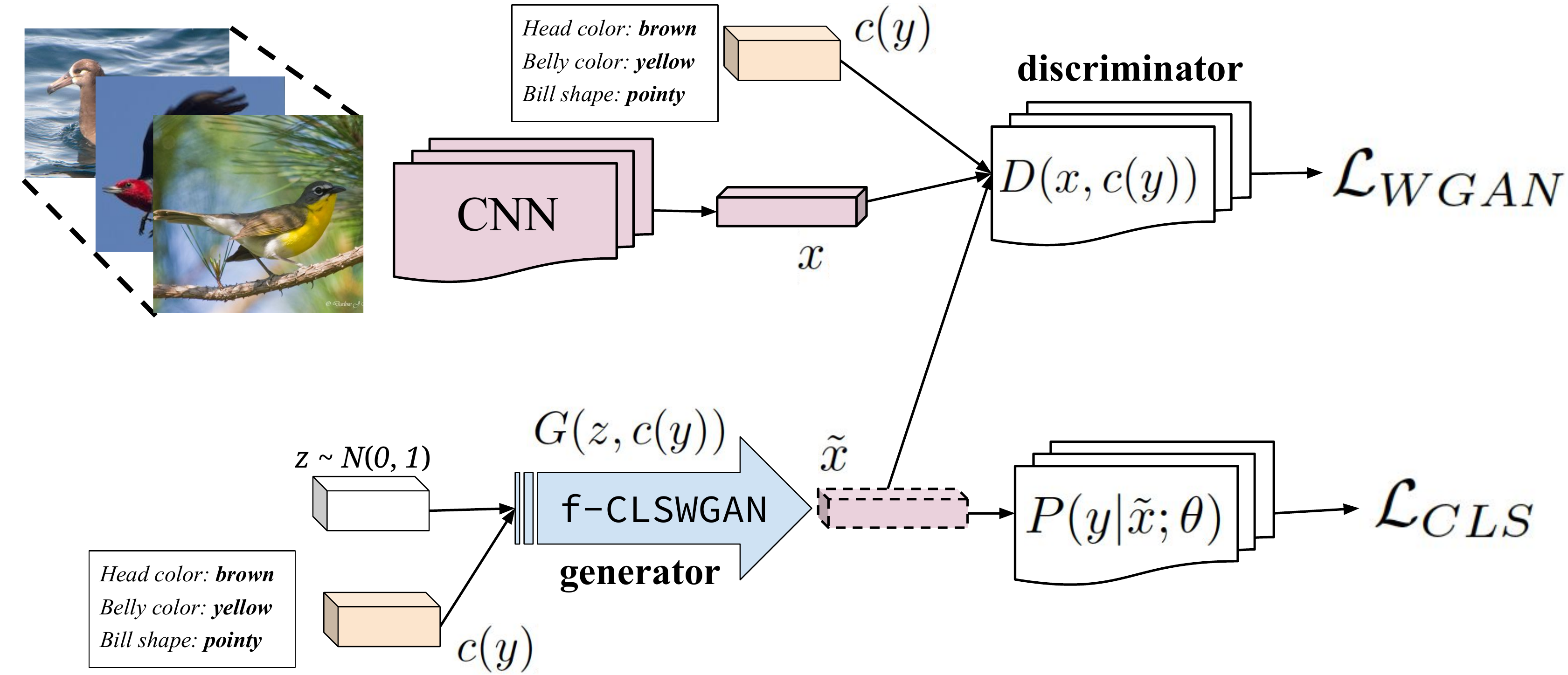}
	\caption{Our \texttt{f-CLSWGAN}: we propose to minimize the classification loss over the generated features and the Wasserstein distance with gradient penalty.}
    \vspace{-3mm}
	\label{fig:teaser}
\end{figure}

\myparagraph{\texttt{f-CLSWGAN}.} \texttt{f-WGAN} does not guarantee that the generated CNN features are well suited for training a discriminative classifier, which is our goal. We conjecture that this issue could be alleviated by encouraging the generator to construct features that can be correctly classified by a discriminative classifier trained on the input data. To this end, we propose to minimize the classification loss over the generated features in our novel \texttt{f-CLSWGAN} formulation. We use the negative log likelihood, 
\begin{align}
\mathcal{L}_{CLS} = -E_{\tilde{x}\sim p_{\tilde{x}}}[\log P(y| \tilde{x}; \theta)], 
\end{align}

where $\tilde{x}=G(z,c(y))$, $y$ is the class label of $\tilde{x}$, $P(y| \tilde{x}; \theta)$ denotes the probability of $\tilde{x}$ being predicted with its true class label $y$. The conditional probability is computed by a linear softmax classifier parameterized by $\theta$, which is pretrained on the real features of seen classes. The classification loss can be thought of as a regularizer enforcing the generator to construct discriminative features. 
Our full objective then becomes,
\begin{align}
\min_G \max_D \mathcal{L}_{WGAN} + \beta \mathcal{L}_{CLS}
\end{align}
where $\beta$ is a hyperparameter weighting the classifier. 

\subsection{Classification}
\label{sec:ALE}

Given $c(u)$ of any unseen class $u\in \mathcal{Y}^u$, by resampling the noise $z$ and then recomputing $\tilde{x} = G(z, c(u))$, arbitrarily many visual CNN features $\tilde{x}$ can be synthesized. After repeating this feature generation process for every unseen class, we obtain a synthetic training set  $\tilde{\mathcal{U}}=\{(\tilde{x}, u, c(u))\}$. We then learn a classifier by training either a multimodal embedding model or a softmax classifier. Our generated features allow to train those methods on the combinations of real seen class data $\mathcal{S}$ and generated unseen class data $\tilde{\mathcal{U}}$.

\myparagraph{\texttt{Multimodal Embedding.}} Many zero-shot learning approaches, e.g. ALE~\cite{APHS15}, DEVISE~\cite{FCSBDRM13}, SJE~\cite{ARWLS15}, ESZSL~\cite{RT15} and LATEM~\cite{XASNHS16}, learn a multimodal embedding between the image feature space $\mathcal{X}$ and the class embedding space $\mathcal{C}$ using seen classes data $\mathcal{S}$. 
With our generated features, those methods can be trained with seen classes data $\mathcal{S}$ together with unseen classes data $\tilde{\mathcal{U}}$ to learn a more robust classifier.
The embedding model $F(x, c(y); W)$, parameterized by $W$, measures the compatibility score between any image feature $x$ and class embedding $c(y)$ pair. Given a query image feature $x$, the classifier searches for the class embedding with the highest compatibility via: 
\begin{equation}
f(x) = \argmax_{y} F(x, c(y); W),
\end{equation}
where in ZSL, $y \in\mathcal{Y}^{u}$ and in GZSL, $y\in\mathcal{Y}^{s} \cup \mathcal{Y}^{u}$.

%
%

\myparagraph{\texttt{Softmax.}} The standard softmax classifier minimizes the negative log likelihood loss,
\begin{align}
\min_\theta  - \frac{1}{|\mathcal{T}|}\sum_{(x,y)\in \mathcal{T}} \log P(y| x; \theta), 
\end{align}
where $\theta\in \mathbb{R}^{d_x \times N}$ is the weight matrix of a fully connected layer which maps the image feature $x$ to $N$ unnormalized probabilities with $N$ being the number of classes, and $P(y| x;\theta) = \frac{\exp(\theta_y^T x)}{\sum_i^N \exp(\theta_i^T x)}$. Depending on the task, $\mathcal{T}=\tilde{\mathcal{U}}$ if it is ZSL and $\mathcal{T}=\mathcal{S} \cup \tilde{\mathcal{U}}$ if it is GZSL. The prediction function is:
\begin{align}
f(x) = \arg\max_{y} P(y| x; \theta),
\end{align}
where in ZSL, $y\in \mathcal{Y}^{u}$ and in GZSL, $y\in \mathcal{Y}^{s} \cup \mathcal{Y}^{u}$.

\section{Experiments}

First we detail our experimental protocol, then we present (1) our results comparing our framework with the state of the art for GZSL and ZSL tasks on four challenging datasets, (2) our analysis of \mthd~\footnote{We denote our \texttt{f-GAN}, \texttt{f-WGAN}, \texttt{f-CLSWGAN} as \texttt{f-xGAN}} under different conditions, (3) our large-scale experiments on ImageNet and (4) our comparison of image and image feature generation.

{
\setlength{\tabcolsep}{4.5pt}
\renewcommand{\arraystretch}{1.2}
\begin{table}[t]
\vspace{-3mm}
 \begin{center}
  \begin{tabular}{ l r c c c c }
    \textbf{Dataset}  & \texttt{att} & \texttt{stc} & $|\mathcal{Y}^{s}|+|\mathcal{Y}^{u}|$ & $|\mathcal{Y}^{s}|$ & $|\mathcal{Y}^{u}|$  \\     \hline
    CUB~\cite{CaltechUCSDBirdsDataset}  & 312 & Y & 200 & 100 + 50 & 50 \\
    FLO~\cite{OxfordFlowersDataset} & -- & Y & $102$ & 62 + 20 & $20$  \\
    SUN~\cite{PH12}  & 102 & N & 717 & 580 + 65 & 72 \\  
    AWA~\cite{LNH13}  & 85 & N & 50 & 27 + 13 & 10
  \end{tabular}
 \end{center}
 \vspace{-4mm}
\caption{CUB, SUN, FLO, AWA datasets, in terms of number of attributes per class (\texttt{att}), sentences (\texttt{stc}), number of classes in training + validation ($\mathcal{Y}^{s}$) and test classes ($\mathcal{Y}^{u}$).}
\label{tab:datasets}
\end{table}
}

{
\setlength{\tabcolsep}{3.5pt}
\renewcommand{\arraystretch}{1.2}
\newcommand{\na}{-}
\newcommand{\phz}{\hphantom{0}}
\begin{table*}[t]
 \centering
 \resizebox{\linewidth}{!}{%
   \begin{tabular}{l l c c c c |c c c |c c c |c c c | c c c  }
     & & \multicolumn{4}{c|}{\textbf{Zero-Shot Learning}} & \multicolumn{12}{c}{\textbf{Generalized Zero-Shot Learning}} \\
     & & \textbf{CUB} & \textbf{FLO} & \textbf{SUN} &  \textbf{AWA} & \multicolumn{3}{c}{\textbf{CUB}} & \multicolumn{3}{c}{\textbf{FLO}} & \multicolumn{3}{c}{\textbf{SUN}} & \multicolumn{3}{c}{\textbf{AWA}}    \\
     Classifier & FG & \textbf{T1} & \textbf{T1} & \textbf{T1} & \textbf{T1}&  \textbf{u} & \textbf{s} & \textbf{H} & \textbf{u} & \textbf{s} & \textbf{H} & \textbf{u} & \textbf{s} & \textbf{H} & \textbf{u} & \textbf{s} & \textbf{H}  \\
     \hline
     \multirow{2}{*}{DEVISE~\cite{FCSBDRM13}} & none &  $52.0$ & $45.9$ & $56.5$ & $54.2$
     & $23.8$ & $53.0$ & $32.8$ & $9.9$ & $44.2$ & $16.2$ &  $16.9$ & $27.4$ & $20.9$ & $13.4$ & $68.7$ & $22.4$  \\
     & \texttt{f-CLSWGAN} & $60.3$ & $60.4$ & $60.9$ & $66.9$ & $52.2$ & $42.4$ & $46.7$ & $45.0$ &  $38.6$& $41.6$ & $38.4$ & $25.4$ & $30.6$ & $35.0$ & $62.8$ & $45.0$ \\
     \hline
     \multirow{2}{*}{SJE~\cite{ARWLS15}} & none &  $53.9$ & $53.4$ & $53.7$ & $65.6$
     & $23.5$ & $59.2$ & $33.6$ & $13.9$ & $47.6$ & $21.5$ &  $14.7$ & $30.5$ & $19.8$ & $11.3$ & $74.6$ & $19.6$ \\
     & \texttt{f-CLSWGAN} & $58.4$ & $67.4$ & $56.5$ & $66.9$ & $48.1$ & $37.4$ & $42.1$ & $52.1$ & $56.2$ & $54.1$ & $36.7$ & $25.0$ & $29.7$ & $37.9$ & $70.1$ & $49.2$ \\
     \hline
     \multirow{2}{*}{LATEM~\cite{XASNHS16}} & none & $49.3$ & $40.4$ & $55.3$ & $55.1$
     & $15.2$ & $57.3$ & $24.0$ & $6.6$ & $47.6$ & $11.5$ &  $14.7$ & $28.8$ & $19.5$ & $7.3$ & $71.7$ & $13.3$  \\
     & \texttt{f-CLSWGAN} & $60.8$ & $60.8$ & $61.3$ & $\mathbf{69.9}$ & $53.6$ & $39.2$ & $45.3$ & $47.2$ & $37.7$ & $41.9$ & $42.4$ & $23.1$ & $29.9$ & $33.0$ & $61.5$ & $43.0$ \\
     \hline
     \multirow{2}{*}{ESZSL~\cite{RT15}} & none &  $53.9$ & $51.0$ & $54.5$ & $58.2$
     & $12.6$ & $63.8$ & $21.0$ & $11.4$ & $56.8$ & $19.0$ &  $11.0$ & $27.9$ & $15.8$ & $6.6$ & $75.6$ & $12.1$  \\
     & \texttt{f-CLSWGAN} & $54.7$ & $54.3$  & $54.0$ & $63.9$ & $36.8$ & $50.9$ & $43.2$ & $25.3$ & $69.2$ & $37.1$ &  $27.8$ & $20.4$ & $23.5$ & $31.1$ & $72.8$ & $43.6$ \\
     \hline 
     \multirow{2}{*}{ALE~\cite{APHS15}} & none & $54.9$ & $48.5$ & $58.1$ & $59.9$
     & $23.7$ & $62.8$ & $34.4$ & $13.3$ & $61.6$ & $21.9$ & $21.8$ & $33.1$ & $26.3$ & $16.8$ & $76.1$ & $27.5$ \\
     & \texttt{f-CLSWGAN} & $\mathbf{61.5}$ & $\mathbf{71.2}$ & $\mathbf{62.1}$ & $68.2$ & $40.2$& $59.3$ & $47.9$ & $54.3$ & $60.3$ & $57.1$ & $41.3$ & $31.1$ & $35.5$ & $47.6$ & $57.2$ & $52.0$ \\
     \hline
     \multirow{2}{*}{\texttt{Softmax}} & none & --  & --  & -- & -- & -- & -- & -- & -- & -- & -- & -- & -- & -- & -- & -- & -- \\
     & \texttt{f-CLSWGAN} & $57.3$ & $67.2$ & $60.8$ & $68.2$ & $43.7$ & $57.7$ & $\mathbf{49.7}$ & $59.0$ & $73.8$ & $\mathbf{65.6}$ &  $42.6$ & $36.6$ & $\mathbf{39.4}$ & $57.9$ & $61.4$ & $\mathbf{59.6}$ \\    
\end{tabular}
   }
\caption{ZSL measuring per-class average Top-1 accuracy (T1) on $\mathcal{Y}^{u}$ and GZSL measuring $\mathbf{u}$ = T1 on $\mathcal{Y}^{u}$, $\mathbf{s}$ = T1 on $\mathcal{Y}^{s}$, H = harmonic mean (FG=feature generator, none: no access to generated CNN features, hence \texttt{softmax} is not applicable). \texttt{f-CLSWGAN} significantly boosts both the ZSL and GZSL accuracy of all classification models on all four datasets. 
}
\label{tab:all}
\end{table*}
}

\myparagraph{Datasets.} Caltech-UCSD-Birds 200-2011 (CUB)~\cite{CaltechUCSDBirdsDataset}, Oxford Flowers (FLO)~\cite{OxfordFlowersDataset} and SUN Attribute (SUN)~\cite{PH12} are all fine-grained datasets. CUB contains 11,788 images from 200 different types of birds annotated with 312 attributes. FLO dataset 8189 images from 102 different types of flowers without attribute annotations. However, for both CUB and FLO we use the fine-grained visual descriptions collected by~\cite{RALS16}. SUN contains 14,340 images from 717 scenes annotated with 102 attributes. Finally,  Animals with Attributes (AWA)~\cite{LNH13} is a coarse-grained dataset with 30,475 images, 50 classes and 85 attributes. Statistics of the datasets are presented in~\autoref{tab:datasets}. We use the zero-shot splits proposed by~\cite{XSA17} for AWA, CUB and SUN insuring that none of the training classes are present in ImageNet~\cite{ImageNet}\footnote{as ImageNet is used for pre-training the ResNet~\cite{HZRS15}}. For FLO, we use the standard split provided by~\cite{RALS16}. 

\myparagraph{Features.}
As real CNN features, we extract $2048$-dim top-layer pooling units of the $101$-layered ResNet~\cite{HZRS15} from the entire image. We do not do any image pre-processing such as cropping or use any other data augmentation techniques. ResNet is pre-trained on ImageNet 1K and not fine-tuned. As synthetic CNN features, we generate $2048$-dim CNN features using our \mthd model. As the class embedding, unless it is stated otherwise, we use per-class attributes for AWA ($85$-dim), CUB ($312$-dim) and SUN ($102$-dim). 
Furthermore, for CUB and Flowers, we extract $1024$-dim character-based CNN-RNN~\cite{RALS16} features from fine-grained visual descriptions (10 sentences per image). None of the $\mathcal{Y}^{u}$ sentences are seen during training the CNN-RNN.
We build per-class sentences by averaging the CNN-RNN features that belong to the same class.

\myparagraph{Evaluation Protocol.}
At test time, in the ZSL setting, the aim is to assign an unseen class label, i.e. $\mathcal{Y}^{u}$ to the test image and in GZSL setting, the search space includes both seen or unseen classes, i.e. $\mathcal{Y}^s \cup \mathcal{Y}^u$.
We use the unified evaluation protocol proposed in~\cite{XSA17}.
In the ZSL setting, the average accuracy is computed independently for each class before dividing their cumulative sum by the number of classes;
i.e., we measure average per-class top-1 accuracy (T1).
In the GZSL setting, we compute average per-class top-1 accuracy on seen classes ($\mathcal{Y}^{s}$) denoted as $\mathbf{s}$, average per-class top-1 accuracy on unseen classes ($\mathcal{Y}^{u}$) denoted as $\mathbf{u}$ and their harmonic mean, i.e. 
%
$H = 2* (\mathbf{s}*\mathbf{u})/(\mathbf{s}+\mathbf{u})$.
%

\myparagraph{Implementation details.} In all \mthd models, both the generator and the discriminator are MLP with LeakyReLU activation. The generator consists of a single hidden layer with 4096 hidden units. Its output layer is ReLU because we aim to learn the top max-pooling units of ResNet-101. While the discriminator of \texttt{f-GAN} has one hidden layer with 1024 hidden units in order to stabilize the GAN training, the discriminators of \texttt{f-WGAN} and \texttt{f-CLSWGAN} have one hidden layer with 4096 hidden units as WGAN~\cite{gulrajani2017improved} does not have instability issues thus a stronger discriminator can be applied here. We do not apply batch normalization our empirical evaluation showed a significant degradation of the accuracy when batch normalization is used. The noise $z$ is drawn from a unit Gaussian with the same dimensionality as the class embedding. We use $\lambda=10$ as suggested in~\cite{gulrajani2017improved} and $\beta=0.01$ across all the datasets.

\begin{figure*}[t]
	\centering
    \begin{subfigure}[]{0.49\linewidth}
        \includegraphics[width=.49\textwidth, trim=10 0 50 0,clip]{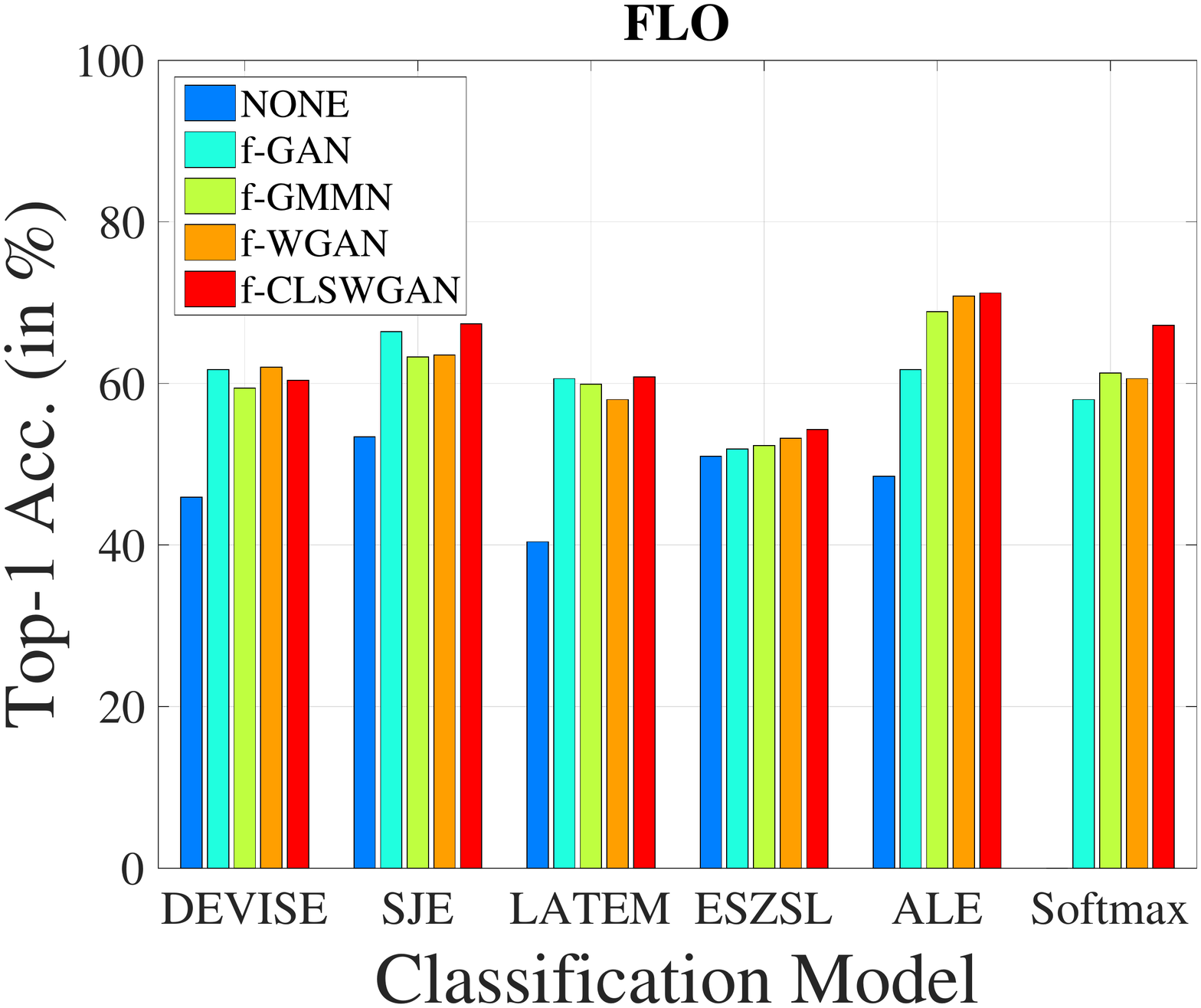}
		\includegraphics[width=.49\textwidth, trim=10 0 50 0,clip]{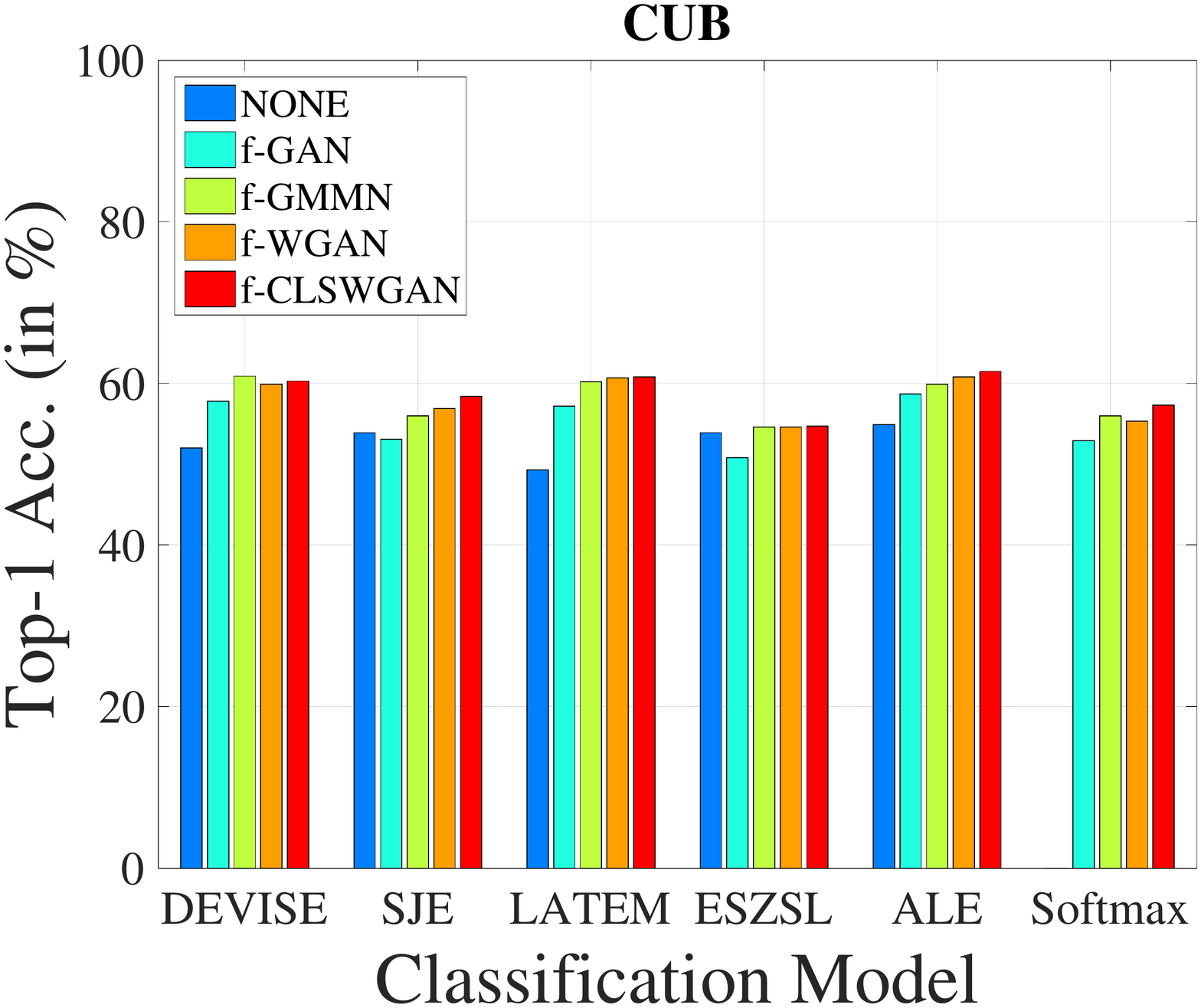}
        \subcaption{Zero-Shot Learning}
        \label{fig:zsl_bar_zsl}
    \end{subfigure}
    \hfill
    \begin{subfigure}[]{0.49\linewidth}
            \includegraphics[width=.49\textwidth, trim=10 0 50 0,clip]{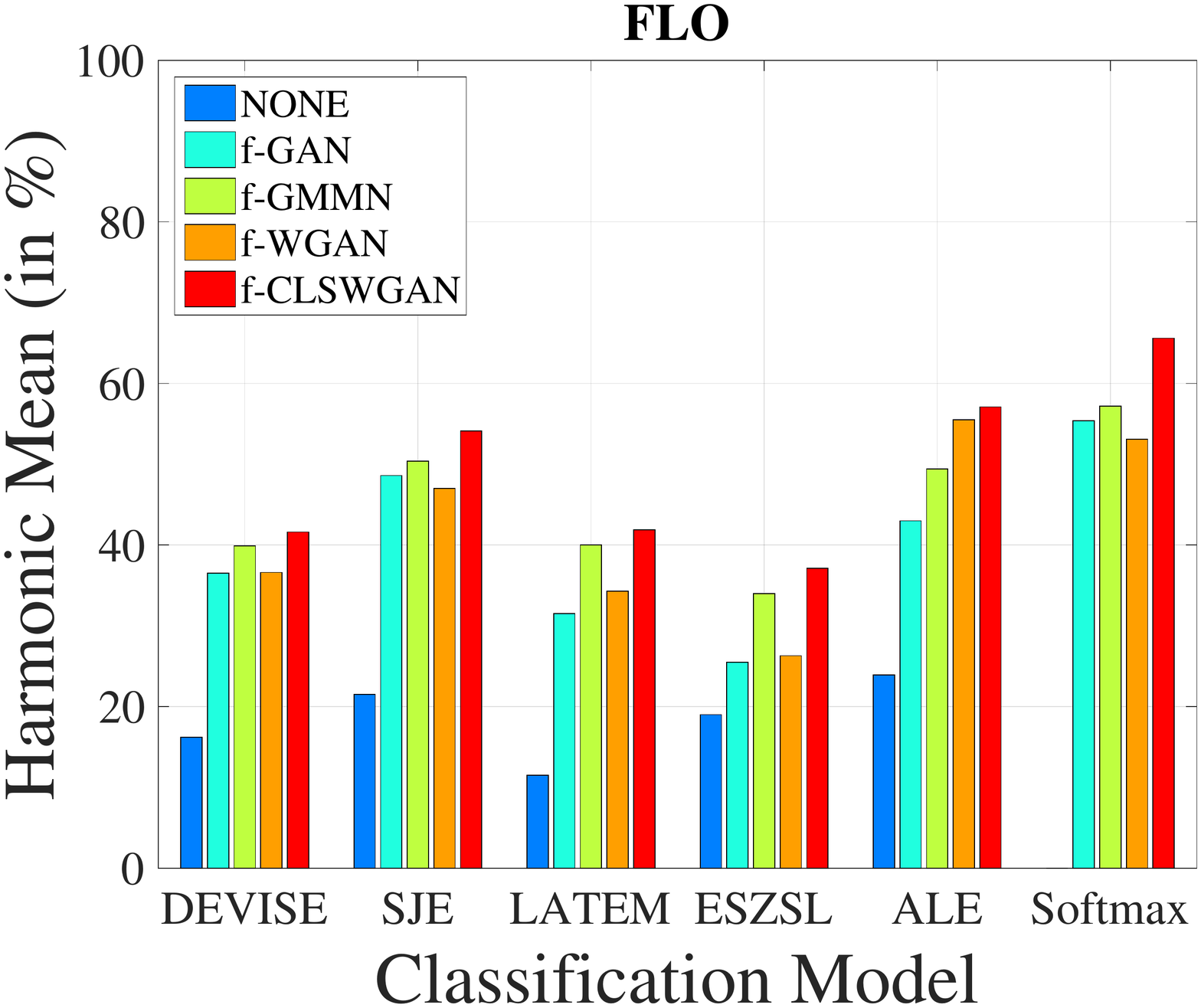}
        \includegraphics[width=.49\textwidth, trim=10 0 50 0,clip]{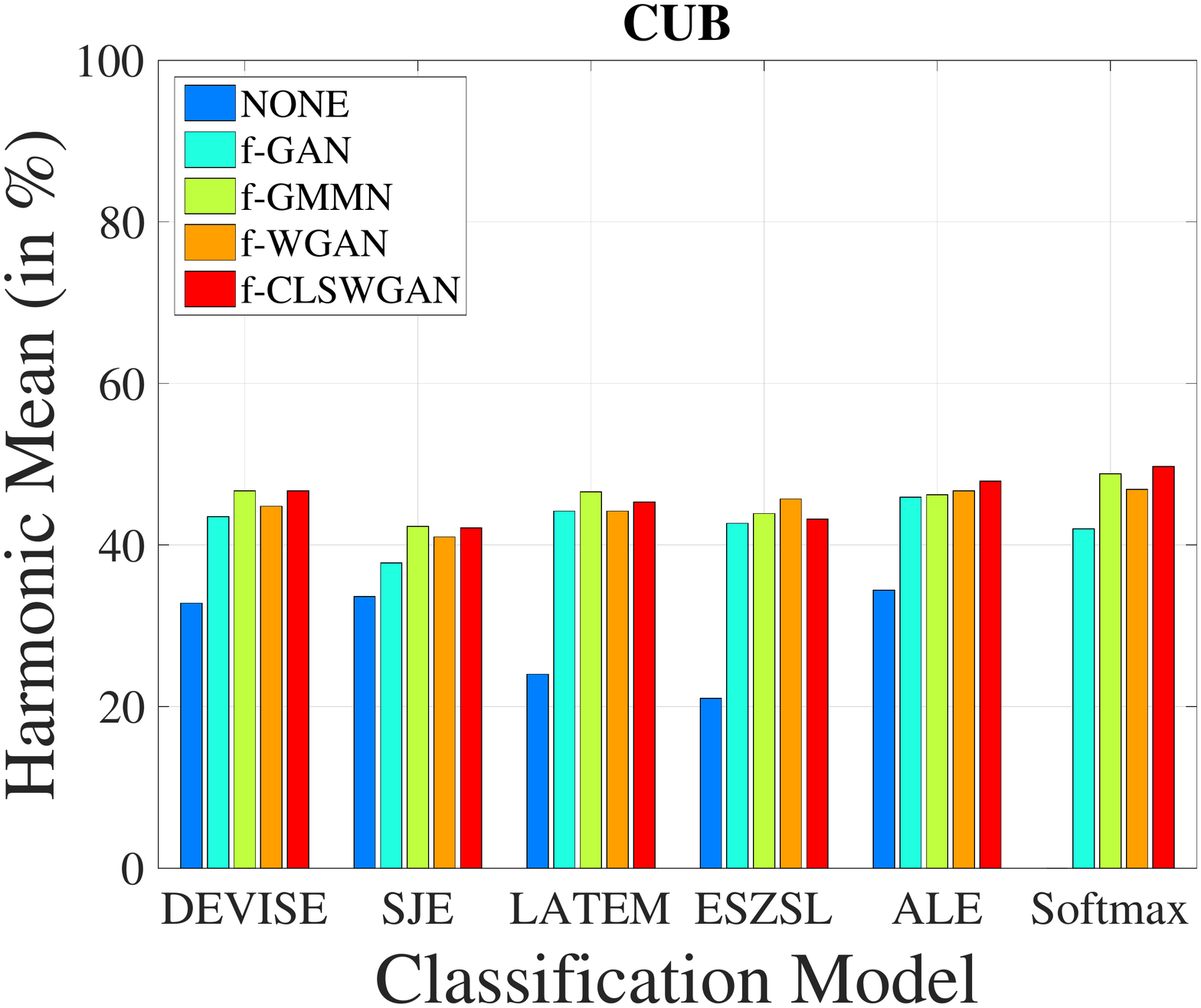}
        \subcaption{Generalized Zero-Shot Learning}
        \label{fig:zsl_bar_gzsl}
    \end{subfigure}
    \vspace{-3mm}
	\caption{Comparing \mthd versions with \texttt{f-GMMN} as well as comparing multimodal embedding methods with softmax.}
	\label{fig:zsl_bar}
\end{figure*}

\subsection{Comparing with State-of-the-Art}

In a first set of experiments, we evaluate our \mthd features in both the ZSL and GZSL settings on four challenging datasets: CUB, FLO, SUN and AWA.
Unless it is stated otherwise, we use \texttt{att} for CUB, SUN, AWA and \texttt{stc} for FLO (as \texttt{att} are not available).
We compare the effect of our feature generating \mthd to $6$ recent state-of-the-art methods~\cite{XSA17}.

\myparagraph{ZSL with \texttt{f-CLSWGAN}.}
We first provide ZSL results with our \texttt{f-CLSWGAN} in \autoref{tab:all} (left).
Here, the test-time search space is restricted to unseen classes $\mathcal{Y}^{u}$.
First, our \texttt{f-CLSWGAN} in all cases improves the state of the art that is obtained without feature generation.
The overall accuracy improvement on CUB is from $54.9\%$ to $61.5\%$, on FLO from $53.4\%$ 
to $71.2\%$, on SUN from $58.1\%$ to $62.1\%$ and on AWA from $65.6\%$ to $69.9\%$, i.e. all quite significant. Another observation is that feature generation is applicable to all the multimodal embedding models and \texttt{softmax}. These results demonstrate that indeed our \texttt{f-CLSWGAN} generates generalizable and strong visual features of previously unseen classes. 

\myparagraph{GZSL with \texttt{f-CLSWGAN}.} Our main interest is GZSL where the test time search space contains both seen and unseen classes, $\mathcal{Y}^{s} \cup \mathcal{Y}^{u}$, and at test time the images come both from seen and unseen classes. Therefore, we evaluate both seen and unseen class accuracy, i.e. $\mathbf{s}$ and $\mathbf{u}$, as well as their harmonic mean (H). The GZSL results with \texttt{f-CLSWGAN} in \autoref{tab:all} (right) demonstrate that for all datasets our \mthd significantly improves the H-measure over the state-of-the-art. On CUB, \texttt{f-CLSWGAN} obtains $49.7\%$ in H measure, significantly improving the state of the art ($34.4\%$), on FLO it achieves $65.6\%$ (vs. $21.9\%$), on SUN it reaches $39.4\%$ (vs. $26.3\%$),
and on AWA it achieves $59.6\%$ (vs. $27.5\%$).
The accuracy boost can be attributed to the strength of the \texttt{f-CLSWGAN} generator learning to imitate CNN features of unseen classes although not having seen any real CNN features of these classes before. 

We also observe that without feature generation on all models the seen class accuracy is significantly higher than unseen class accuracy, which indicates that many samples are incorrectly assigned to one of the seen classes. Feature generation through \texttt{f-CLSWGAN} finds a balance between seen and unseen class accuracies by improving the unseen class accuracy while maintaining the accuracy on seen classes. Furthermore, we would like to emphasize that the simple \texttt{softmax} classifier beats all the models and is now applicable to GZSL thanks to our CNN feature generation. This shows the true potential and generalizability of feature generation to various tasks.

\myparagraph{ZSL and GZSL with \mthd.} The generative model is an important component of our framework. Here, we evaluate all versions of our \mthd and \texttt{f-GMMN} for it being a strong alternative. We show ZSL and GZSL results of all classification models in \autoref{fig:zsl_bar}. We selected CUB and FLO for them being fine-grained datasets, however we provide full numerical results and plots in the supplementary which shows that our observations hold across datasets. Our first observation is that for both ZSL and GZSL settings all generative models improve in all cases over ``none'' with no access to the synthetic CNN features. This applies to the GZSL setting and the difference between ``none'' and \mthd is strikingly significant. Our second observation is that our novel \texttt{f-CLSWGAN} model is the best performing generative model in almost all cases for both datasets. Our final observation is that although \texttt{f-WGAN} rarely performs lower than \texttt{f-GMMN}, e.g. ESZL on FLO, our \texttt{f-CLSWGAN} which uses a classification loss in the generator recovers from it and achieves the best result among all these generative models. We conclude from these experiments that generating CNN features to support the classifier when there is missing data is a technique that is flexible and strong.

We notice that recently~\cite{zhang2016learning} has shown great performance on the old splits of AWA and CUB datasets. We compare our method with~\cite{zhang2016learning} using the same evaluation protocol as our paper, i.e same data splits and evaluation metrics. On AWA, in ZSL task, the comparison is 66.1\% vs 69.9\% (ours) and in GZSL task,  it is 41.4\% vs 59.6\% (ours). On CUB, in ZSL task, the comparison is 50.1\% vs 61.5\% (ours) and in GZSL task it is 29.2\% vs 49.7\% (ours). 

\begin{figure}[t]
	\centering
        \includegraphics[width=.23\textwidth, trim=0 0 70 0,clip]{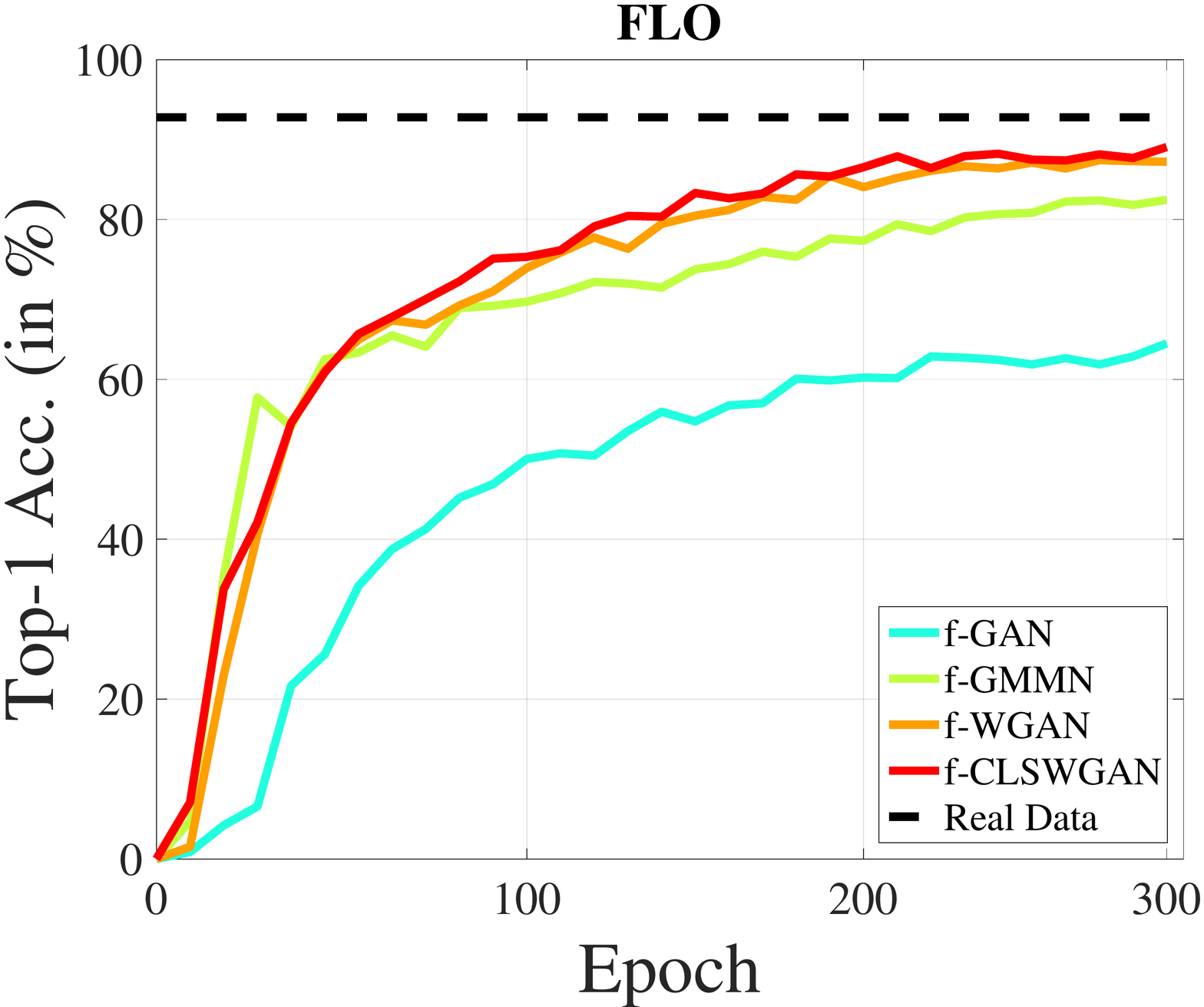} 
		\includegraphics[width=.23\textwidth, trim=0 0 70 0,clip]{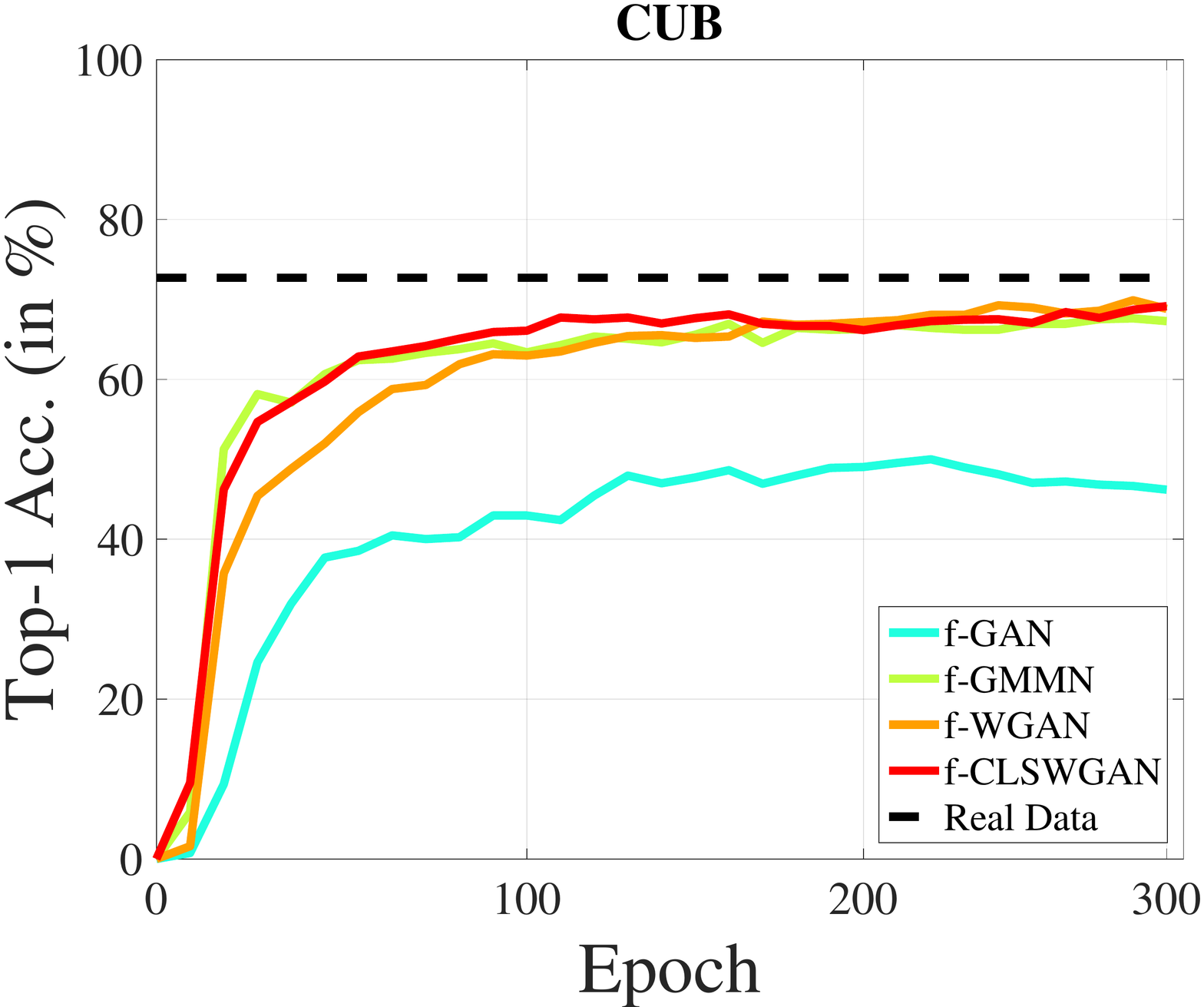}       
    	\caption{Measuring the seen class accuracy of the classifier trained on generated features of seen classes w.r.t. the training epochs (with \texttt{softmax}).} 
    \vspace{-3mm}
	\label{fig:gan_seen_acc}
\end{figure}

\subsection{Analyzing \mthd Under Different Conditions}
In this section, we analyze \mthd in terms of stability, generalization, CNN architecture used to extract real CNN features and the effect of class embeddings on two fine-grained datasets, namely CUB and FLO. 

\myparagraph{Stability and Generalization.} We first analyze how well different generative models fit the seen class data used for training. Instead of using Parzen window-based log-likelihood~\cite{GPMXWDOCB14} that is unstable, we train a softmax classifier with generated features of seen classes and report the classification accuracy on a held-out test set. \autoref{fig:gan_seen_acc} shows the classification accuracy w.r.t the number of training epochs. On both datasets, we observe a stable training trend. On FLO, compared to the supervised classification accuracy obtained with real images, i.e. the upper bound marked with dashed line, \texttt{f-GAN} remains quite weak even after convergence, which indicates that \texttt{f-GAN} has underfitting issues. A strong alternative is \texttt{f-GMMN} leads to a significant accuracy boost while our \texttt{f-WGAN} and \texttt{f-CLSWGAN} improve over \texttt{f-GMMN} and almost reach the supervised upper bound. 

\begin{figure}[t]
	\centering
            \includegraphics[width=.23\textwidth, trim=10 0 50 0,clip]{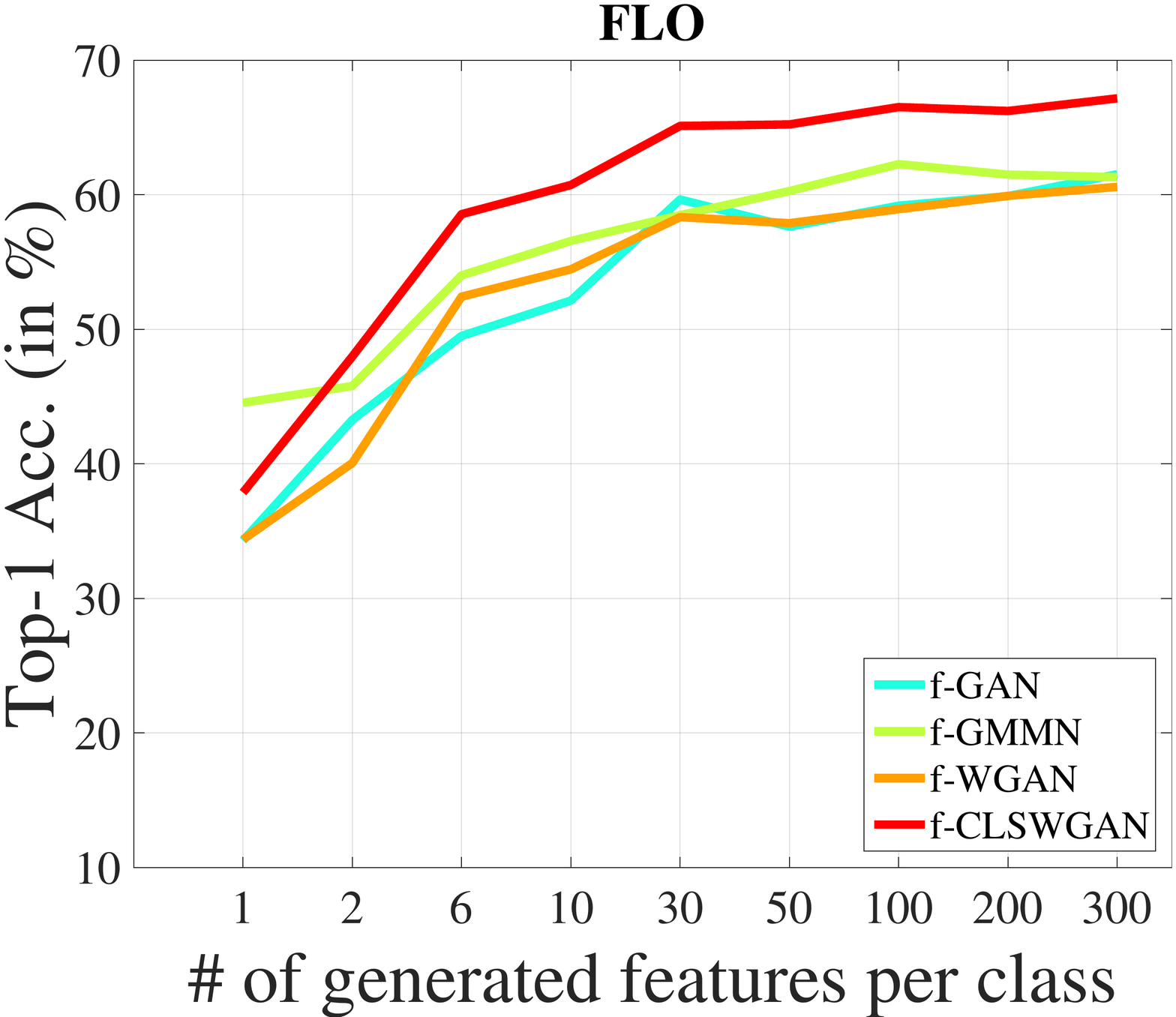}
		\includegraphics[width=.23\textwidth, trim=10 0 50 0,clip]{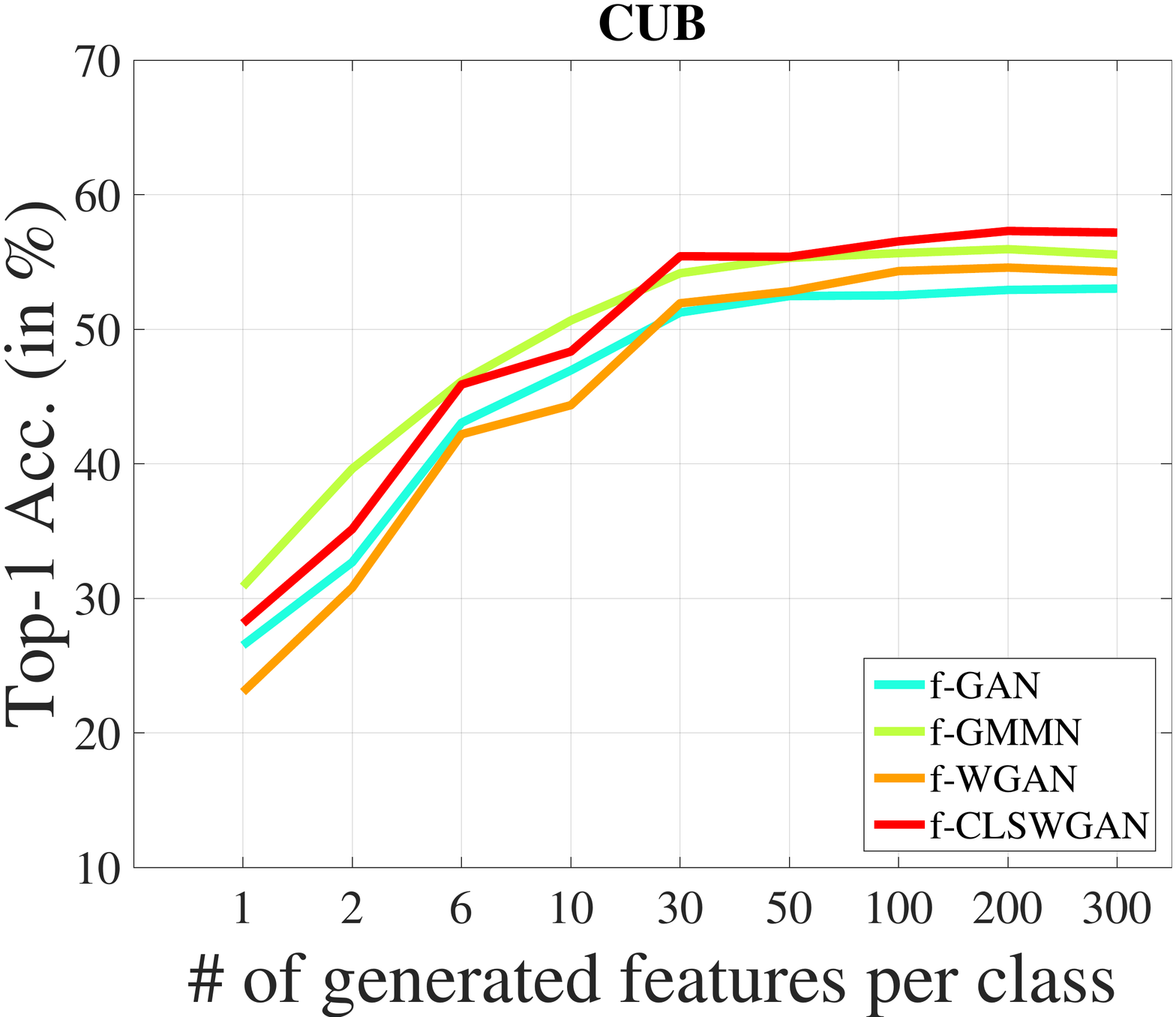}
	\caption{Increasing the number of generated \mthd features wrt unseen class accuracy (with \texttt{softmax}) in ZSL.}
	\label{fig:unseen_number}
\end{figure}

After having established that our \mthd leads to a stable training performance and generating highly descriptive features, 
we evaluate the generalization ability of the \mthd generator to unseen classes. Using the pre-trained model, we generate CNN features of unseen classes. We then train a softmax classifier using these synthetic CNN features of unseen classes with real CNN features of seen classes. On the GZSL task, \autoref{fig:unseen_number} shows that increasing the number of generated features of unseen classes from 1 to 100 leads to a significant boost of accuracy, e.g. $28.2\%$ to $56.5\%$ on CUB and $37.9\%$ to $66.5\%$ on FLO. As in the case for generating seen class features, here the ordering is \texttt{f-GAN} $<$ \texttt{f-WGAN} $<$ \texttt{f-GMMN} $<$ \texttt{f-CLSWGAN} on CUB and \texttt{f-GAN} $<$ \texttt{f-GMMN} $<$ \texttt{f-WGAN} $<$ \texttt{f-CLSWGAN} on FLO. With these results, we argue that if the generative model can generalize well to previously unseen data distributions, e.g. perform well on GZSL task, they have practical use in a wide range of real-world applications. Hence, we propose to quantitatively evaluate the performance of generative models on the GZSL task. 

{
\setlength{\tabcolsep}{5pt}
\renewcommand{\arraystretch}{1.2}
\begin{table}[t]
 \centering
   \begin{tabular}{l l c c c c }
   CNN & FG & $\mathbf{u}$ & $\mathbf{s}$ & \textbf{H} \\ 
   \hline
   \multirow{2}{*}{GoogLeNet} & none  & $20.2$ & $35.7$ & $25.8$ \\ 
    & \texttt{f-CLSWGAN}  &  $35.3$ & $38.7$ & $36.9$   \\ 
    \hline  
  \multirow{2}{*}{ResNet-101} & none  & $23.7$ & $62.8$ & $34.4$ \\ 
   & \texttt{f-CLSWGAN}  &  $43.7$ & $57.7$ & $49.7$   \\ 
   \end{tabular} 
\caption{GZSL results with GoogLeNet vs ResNet-101 features on CUB (CNN: Deep Feature Encoder Network, FG: Feature Generator, $\mathbf{u}$ = T1 on $\mathcal{Y}^{u}$, $\mathbf{s}$ = T1 on $\mathcal{Y}^{s}$, H = harmonic mean, ``none''= no generated features).
}
\label{tab:res_goog}
\end{table}
}

\myparagraph{Effect of CNN Architectures.} The aim of this study is to determine the effect of the deep CNN encoder that provides real features to our \mthd discriminator. In~\autoref{tab:res_goog}, we first observe that with GoogLeNet features, the results are lower compared to the ones obtained with ResNet features.
This indicates that ResNet features are stronger than GoogLeNet, which is expected. Besides, most importantly, with both CNN architectures we observe that our \mthd outperforms the ``none'' by a large margin. Specifically, the accuracy increases from $25.8\%$ to $36.9\%$ for GoogleNet features and $34.4\%$ to $49.7\%$ for ResNet features. Those results are encouraging as they demonstrate that our \mthd is not limited to learning the distribution of ResNet-101 features, but also able to learn other feature distributions.

{
\setlength{\tabcolsep}{5pt}
\renewcommand{\arraystretch}{1.2}
\begin{table}[t]
 \centering
   \begin{tabular}{l l c c c }
     C & FG & $\mathbf{u}$ & $\mathbf{s}$ & \textbf{H} \\ 
     \hline  
     \multirow{2}{*}{Attribute (\texttt{att})}& none  &  $23.7$ & $62.8$ &  $34.4$ \\ 
      & \texttt{f-CLSWGAN} & $43.7$ & $57.7$ & $49.7$ \\ 
      \hline
     \multirow{2}{*}{Sentence (\texttt{stc})} & none   & $38.8$ & $53.8$ & $45.1$    \\ 
      & \texttt{f-CLSWGAN} & $50.3$ & $58.3$ & $54.0$   \\ 
   \end{tabular} 
\caption{GZSL results with conditioning \texttt{\mthd} with \texttt{stc} and \texttt{att} on CUB (C: Class embedding, FG: Feature Generator,  $\mathbf{u}$ = T1 on $\mathcal{Y}^{u}$, $\mathbf{s}$ = T1 on $\mathcal{Y}^{s}$, H = harmonic mean, ``none''= no generated features).}
\label{tab:language}
\end{table}
}

\myparagraph{Effect of Class Embeddings.} The conditioning variable, i.e. class embedding, is an important component of our \mthd. Therefore, we evaluate two different class embeddings, per-class attributes (\texttt{att}) and per-class sentences (\texttt{stc}) on CUB as this is the only dataset that has both. In~\autoref{tab:language}, we first observe that \texttt{f-CLSWGAN} features generated with \texttt{att} not only lead to a significantly higher result ($49.7\%$ vs $34.4\%$),  $\mathbf{s}$ and $\mathbf{u}$ are much more balanced ($57.7\%$ and $43.7\%$ vs. $62.8\%$ and $23.7\%$) compared to the state-of-the-art, i.e. ``none''. This is because generated CNN features help us explore the space of unseen classes whereas the state of the art learns to project images closer to seen class embeddings.

Finally, \texttt{f-CLSWGAN} features generated with per-class \texttt{stc} significantly improve results over \texttt{att}, achieving $54.0\%$ in H measure,
and also leads to a notable $\mathbf{u}$ of $50.3\%$ without hurting $\mathbf{s}$ ($58.3\%$).
This is due to the fact that \texttt{stc} leads to high quality features~\cite{RALS16} reflecting the highly descriptive semantic content language entails and it shows that our \texttt{f-CLSWGAN} is able to learn higher quality CNN features given a higher quality conditioning signal.

\subsection{Large-Scale Experiments}
\begin{figure}[t]
	\centering
		\includegraphics[width=.48\columnwidth, trim=10 10 50 0,clip]{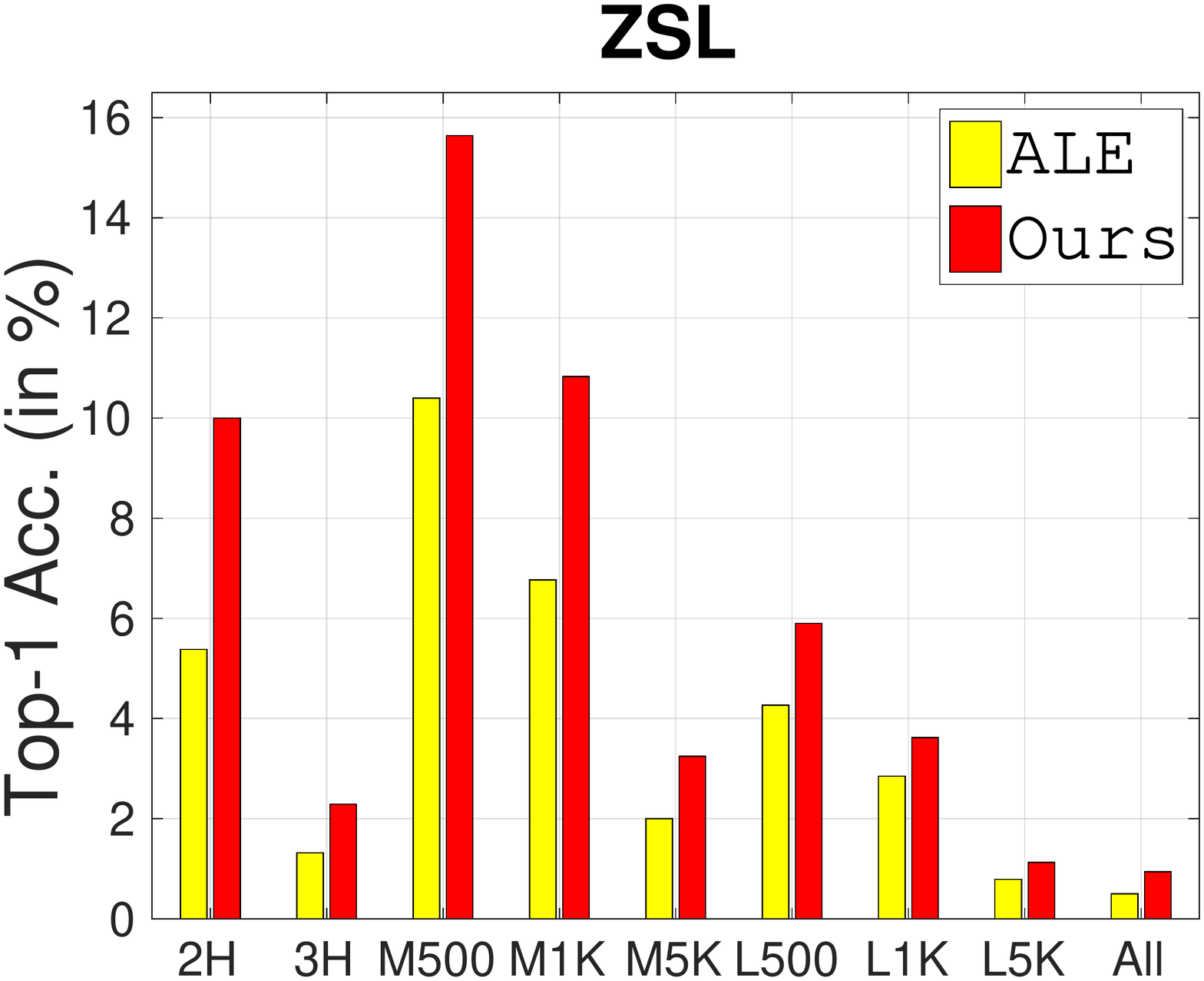}
        \includegraphics[width=.48\columnwidth, trim=10 10 50 0,clip]{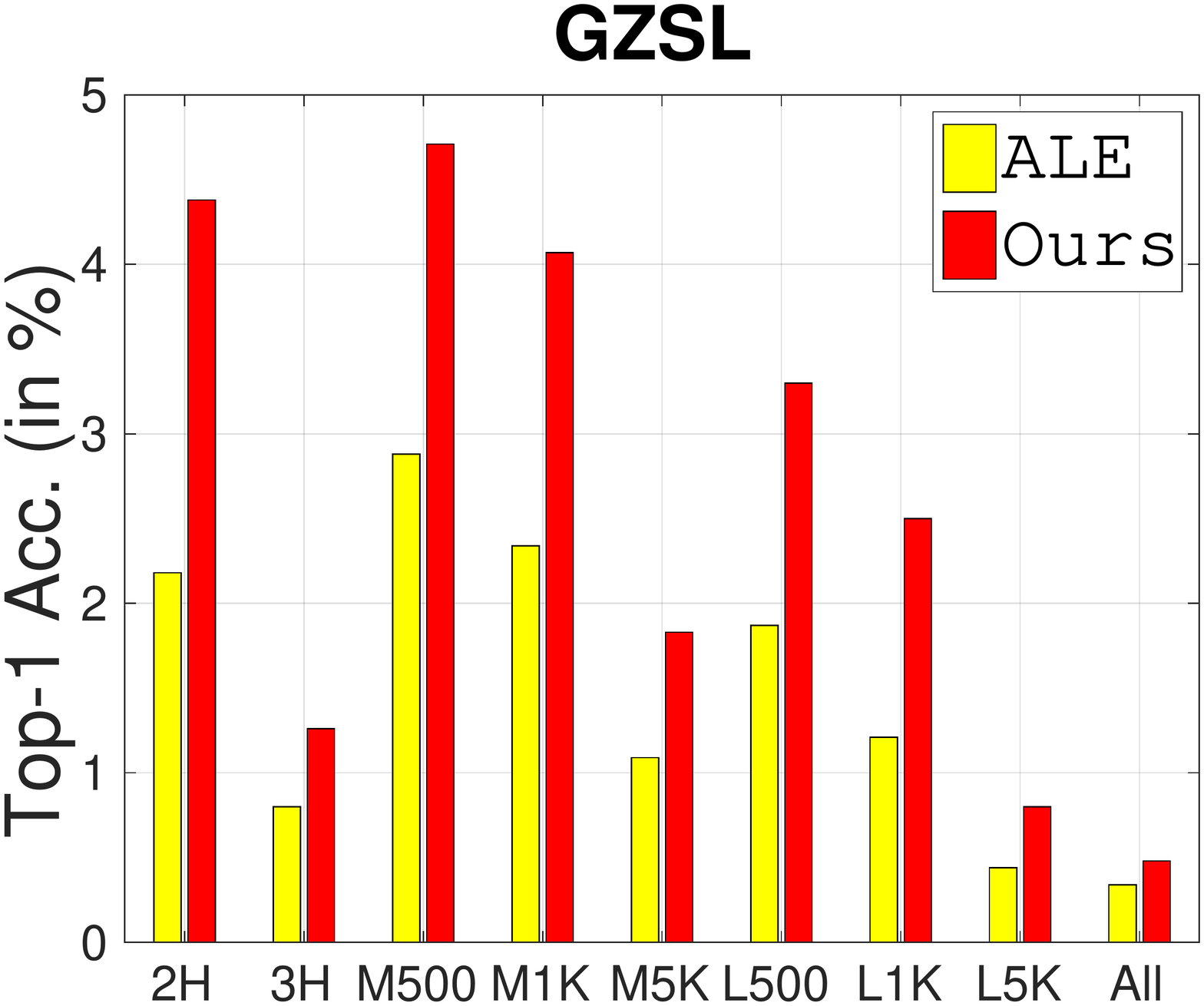}
	\caption{ZSL and GZSL results on ImageNet (ZSL: T1 on $\mathcal{Y}^{u}$, GZSL: T1 on $\mathcal{Y}^{u}$). The splits, ResNet features and Word2Vec are provided by~\cite{XSA17}. ``Ours'' = feature generator: \texttt{f-CLSWGAN}, classifier: \texttt{softmax}.}
	\label{fig:imagenet}
\end{figure}

Our large-scale experiments follow the same zero-shot data splits of~\cite{XSA17} and serve two purposes.
First, we show the generalizability of our approach by conducting ZSL and GZSL experiments on ImageNet~\cite{ImageNet} for it being the largest-scale single-label image dataset, i.e. with 21K classes and 14M images. Second, as ImageNet does not contain \texttt{att}, we use as a (weak) conditioning signal Word2Vec~\cite{MSCCD13} to generate \texttt{f-CLSWGAN} features. \autoref{fig:imagenet} shows that \texttt{softmax} as a classifier obtains the state-of-the-art of ZSL and GZSL on ImageNet, significantly improving over ALE~\cite{APHS15}. These results show that our \texttt{f-CLSWGAN} is able to generate high quality CNN features also with Word2Vec as the class embedding. 

For ZSL, for instance, with the 2H split ``Ours'' almost doubles the performance of ALE ($5.38\%$ to $10.00\%$) and in one of the extreme cases, e.g. with L1K split, the accuracy improves from $2.85\%$ to $3.62\%$. For GZSL the same observations hold, i.e. the gap between ALE and ``Ours'' is $2.18$ vs $4.38$ with 2H split and $1.21$ vs $2.50$ with L1K split. Note that, \cite{XSA17} reports the highest results with SYNC~\cite{CCGS16} and ``Ours'' improves over SYNC as well, e.g. $9.26\%$ vs $10.00\%$ with 2H and $3.23\%$ vs $3.56\%$ with L1K. With these results we emphasize that with a supervision as weak as a Word2Vec signal, our model is able to generate CNN features of unseen classes and operate at the ImageNet scale. This does not only hold for the ZSL setting which discards all the seen classes from the test-time search space assuming that the evaluated images will belong to one of the unseen classes. It also holds for the GZSL setting where no such assumption has been made. Our model generalizes to previously unseen classes even when the seen classes are included in the search space which is the most realistic setting for image classification.

\subsection{Feature vs Image Generation}

As our main goal is solving the GZSL task which suffers from the lack of visual training examples, one naturally thinks that image generation serves the same purpose. Therefore, here we compare generating images and image features for the task of GZSL. We use the StackGAN~\cite{han2017stackgan} to generate $256\times 256$ images conditioned on sentences. 


{
\setlength{\tabcolsep}{3.5pt}
\renewcommand{\arraystretch}{1.2}
\begin{table}[t]
 \centering
   \begin{tabular}{l c c c c c c}
   & \multicolumn{3}{c}{$\mathbf{CUB}$} & \multicolumn{3}{c}{$\mathbf{FLO}$} \\
   Generated Data & $\mathbf{u}$ & $\mathbf{s}$ & \textbf{H} & $\mathbf{u}$ & $\mathbf{s}$ & \textbf{H} \\ 
   \hline
   none & $38.8$ & $53.8$ & $45.1$ & $13.3$ & $61.6$ & $21.9$ \\
   Image (with \cite{han2017stackgan}) & $23.8$ & $48.5$ & $31.9$ & $39.4$ & $64.9$ & $49.0$ \\
   CNN feature (Ours) & $50.3$ & $58.3$ & $\mathbf{54.0}$ & $59.0$ & $73.8$ & $\mathbf{65.6}$ \\
   \end{tabular} 
\caption{Summary Table ($\mathbf{u}$ = T1 on $\mathcal{Y}^{u}$, $\mathbf{s}$ = T1 accuracy on $\mathcal{Y}^{s}$, H = harmonic mean, class embedding = \texttt{stc}). ``none'': ALE with no generated features. 
}
\label{tab:imggen}
\end{table}
}%

In \autoref{tab:imggen}, we compare GZSL results obtained with ``none'', i.e. with an ALE model trained on real images of seen classes, Image, i.e. image features extracted from $256\times 256$ synthetic images generated by StackGAN~\cite{han2017stackgan} and CNN feature, i.e. generated by our \texttt{f-CLSWGAN}. 

Between ``none'' and ``Image'', we observe that generating images of unseen classes improves the performance i.e. harmonic mean on FLO ($49.0\%$ for ``Image'' vs $21.9\%$ for ``none''), but hurts the performance on CUB ($31.9\%$ for ``Image'' vs $45.1\%$ for ``none''). This is because generating birds is a much harder task than generating flowers.
Upon visual inspection, we have observed that although many images have an accurate visual appearance as birds or flowers, they lack the necessary discriminative details to be classified correctly and the generated images are not class-consistent. On the other hand, generating CNN features leads to a significant boost of accuracy, e.g. $54.0\%$ on CUB and $65.6\%$ on FLO which is clearly higher than having no generation, i.e. ``none'', and image generation. 

We argue that image feature generation has the following advantages. First, the number of generated image features is limitless. Second, the image feature generation learns from compact invariant representations obtained by a deep network trained on a large-scale dataset such as ImageNet, therefore the feature generative network can be quite shallow and hence computationally efficient.
Third, generated CNN features are highly discriminative, i.e. they lead to a significant boost in performance of both ZSL and GZSL. Finally, image feature generation is a much easier task as the generated data is much lower dimensional than high quality images necessary for discrimination.


\section{Conclusion}

In this work, we propose \mthd, a learning framework for feature generation followed by classification, to tackle the generalized zero-shot learning task.
Our \mthd model adapts the conditional GAN architecture that is frequently used for generating image pixels to generate CNN features.
In \texttt{f-CLSWGAN}, we improve WGAN by adding a classification loss on top of the generator, enforcing it to generate features that are better suited for classification. 
In our experiments, we have shown that generating features of unseen classes allows us to effectively use softmax classifiers for the GZSL task.

Our framework is generalizable as it can be integrated to various deep CNN architectures, i.e. GoogleNet and ResNet as a pair of the most widely used architectures. It can also be deployed with various classifiers, e.g. ALE, SJE, DEVISE, LATEM, ESZSL that constitute the state of the art for ZSL but also the GZSL accuracy improvements obtained with softmax is important as it is a simple classifier that could not be used for GZSL before this work. Moreover, our features can be generated via different sources of class embeddings, e.g. Sentence, Attribute, Word2vec, and applied to different datasets, i.e. CUB, FLO, SUN, AWA being fine and coarse-grained ZSL datasets and ImageNet being a truly large-scale dataset.

Finally, based on the success of our framework, we motivated the use of GZSL tasks as an auxiliary method for evaluation of the expressive power of generative models in addition to manual inspection of generated image pixels which is tedious and prone to errors. For instance, WGAN~\cite{gulrajani2017improved} has been proposed and accepted as an improvement over GAN~\cite{GPMXWDOCB14}. This claim is supported with evaluations based on manual inspection of the images and the inception score. Our observations in~\autoref{fig:zsl_bar} and in~\autoref{fig:unseen_number} support this and follow the same ordering of the models, i.e. WGAN improves over GAN in ZSL and GZSL tasks. Hence, while not being the primary focus of this paper, we strongly argue, that ZSL and GZSL are suited well as a testbed for comparing generative models.


{\small
\bibliographystyle{ieee}
\bibliography{egbib}
}

\end{document}